%% file: neurips_2026.tex
\documentclass{article}

    \PassOptionsToPackage{numbers, compress}{natbib}
\usepackage[preprint]{neurips_2026}

\usepackage[utf8]{inputenc} % allow utf-8 input
\usepackage[T1]{fontenc}    % use 8-bit T1 fonts
\usepackage{hyperref}       % hyperlinks
\usepackage{url}            % simple URL typesetting
\usepackage{booktabs}       % professional-quality tables
\usepackage{amsfonts}       % blackboard math symbols
\usepackage{nicefrac}       % compact symbols for 1/2, etc.
\usepackage{microtype}      % microtypography
\usepackage{xcolor}         % colors
\usepackage{graphicx}       % images
\usepackage{colortbl}
\usepackage{amsmath}
\usepackage{amssymb}
\usepackage{wrapfig}        % wrap text around tables/figures
\usepackage{algorithm}      % algorithm float
\usepackage{algpseudocode}  % algorithmic pseudocode

\title{Ouroboros-Spatial: Closing the Data-Model Loop for Spatial Reasoning}

\author{%
\begin{tabular}{ccc}
\textbf{Enhan Zhao}$^{1,2}$ &
\textbf{Wei Wu}$^{2,\dag}$ &
\textbf{Yuanrui Zhang}$^{1}$ \\[0.25em]
\multicolumn{3}{c}{
  \textbf{Xueliang Zhao}$^{3}$
  \qquad
  \textbf{Di He}$^{1,\dag}$
}
\end{tabular}
\\[0.7em]
$^{1}$Peking University
\\
$^{2}$Ant International
\\
$^{3}$The University of Hong Kong
\\
$^{\dag}$Corresponding authors.
\\[0.4em]
\texttt{\{morrezhao,yuanruizhang25\}@stu.pku.edu.cn}
\\
\texttt{di\_he@pku.edu.cn}
\\
\texttt{\{wuwei19850318,zhaoxlpku\}@gmail.com}
}

\begin{document}

\maketitle

\input{chap/00-abs}

\input{chap/01-intro}

\input{chap/02-related_work}

\input{chap/03-method}

\input{chap/04-experiment}

\input{chap/05-conclusion}

% \begin{ack}
% Use unnumbered first level headings for the acknowledgments. All acknowledgments
% go at the end of the paper before the list of references. Moreover, you are required to declare
% funding (financial activities supporting the submitted work) and competing interests (related financial activities outside the submitted work).
% More information about this disclosure can be found at: \url{https://neurips.cc/Conferences/2026/PaperInformation/FundingDisclosure}.

% Do {\bf not} include this section in the anonymized submission, only in the final paper. You can use the \texttt{ack} environment provided in the style file to automatically hide this section in the anonymized submission.
% \end{ack}

\bibliographystyle{plainnat}
\bibliography{references}

%%%%%%%%%%%%%%%%%%%%%%%%%%%%%%%%%%%%%%%%%%%%%%%%%%%%%%%%%%%%

\appendix

\input{chap/06-appendix}

%%%%%%%%%%%%%%%%%%%%%%%%%%%%%%%%%%%%%%%%%%%%%%%%%%%%%%%%%%%%

% \newpage
% \input{checklist.tex}

\end{document}

%% file: chap/00-abs.tex
% Abstract for: Ouroboros-Spatial: Closing the Data-Model Loop for Spatial Reasoning
% NeurIPS 2026 submission

\begin{abstract}
Spatial reasoning remains a persistent challenge for multimodal large language models (MLLMs). Existing approaches largely rely on large-scale, statically curated datasets, where all training samples are treated uniformly regardless of the model's evolving capabilities. This static paradigm is inherently data-inefficient: training capacity is often spent on samples that are either trivial or overly difficult for the model at its current stage. To address this limitation, we propose \textbf{Ouroboros-Spatial}, a self-evolving training framework in which the model plays dual roles as a \emph{proposer} and a \emph{solver}. In each iteration, a frozen proposer generates spatial question-answer (QA) pairs from 3D scene metadata and raw video frames, together with executable code for deriving reliable ground truth. A learnable solver is then fine-tuned on the accepted samples, and its per-sample prediction confidence is used as a difficulty signal. This signal is fed back to the proposer in the next iteration, guiding it to generate questions better matched to the solver’s current capabilities. Through this closed-loop design, the training distribution co-evolves with model ability, reducing redundant trivial examples while filtering out ambiguous or uninformative samples with limited learning value. Across six spatial reasoning benchmarks, Ouroboros-Spatial substantially improves Qwen3-VL-4B and Qwen3-VL-8B while using \emph{an order of magnitude fewer} training examples than recent large-scale curated datasets. On VSI-Bench, it yields absolute gains of 9.9 and 6.8 points for the 4B and 8B models, respectively, enabling both to outperform a wide range of strong open-source and proprietary baselines.
\end{abstract}

%% file: chap/01-intro.tex
% Introduction for: Ouroboros-Spatial: Closing the Data-Model Loop for Spatial Reasoning
% NeurIPS 2026 submission

\section{Introduction}
\label{sec:intro}
The rapid advancement of multimodal foundation models has substantially expanded the frontier of machine intelligence, shifting reasoning from operating primarily in the symbolic space toward integrated cross-modal understanding and analysis that jointly leverage visual and textual signals \cite{team2026kimi,Qwen35blog}. However, despite their strong performance on relatively basic tasks such as question-answering over images and figures \cite{yue2025mmmu,wang2024charxiv,wang2024measuring}, state-of-the-art multimodal large language models (MLLMs) still struggle with tasks requiring 3D geometric structure inference and complex spatiotemporal relationship understanding. Yet, these capabilities are critical for a wide range of real-world applications, including autonomous driving, robotics, embodied intelligence, and many other domains.

An important lesson from large language models (LLMs) is that reasoning capabilities can be continuously improved through data scaling, including both challenging problems and chain-of-thought trajectories \cite{zhao2025promptcot2,moshkov2025aimo}. Motivated by this observation, the research community has begun to scale up training data for spatial reasoning and investigate whether similar gains can be achieved in this domain \cite{feng2025vica,yang2024cambrians,sensenova2024,sun2025spacevista,fan2025vlm3r}. For example, a recent study \cite{yang2024cambrians} leveraged annotated, simulated, and unannotated visual data, together with carefully designed templates, to construct a large-scale instruction-tuning dataset containing $590$k examples. Through pure supervised fine-tuning, this dataset delivered more than $30$\% absolute improvement over the base models. Despite these promising results, existing spatial-reasoning studies for MLLMs typically rely on fixed data-generation pipelines, leaving the reasoning capabilities of MLLMs constrained by the static templates or prompts used during question generation. More importantly, because data curation and model optimization are treated as two disentangled stages, it remains unclear which types of data are most effective at different phases of training, making it difficult to develop a cost-effective optimization recipe.

\begin{figure}[t]
    \centering
    \includegraphics[width=0.85\linewidth]{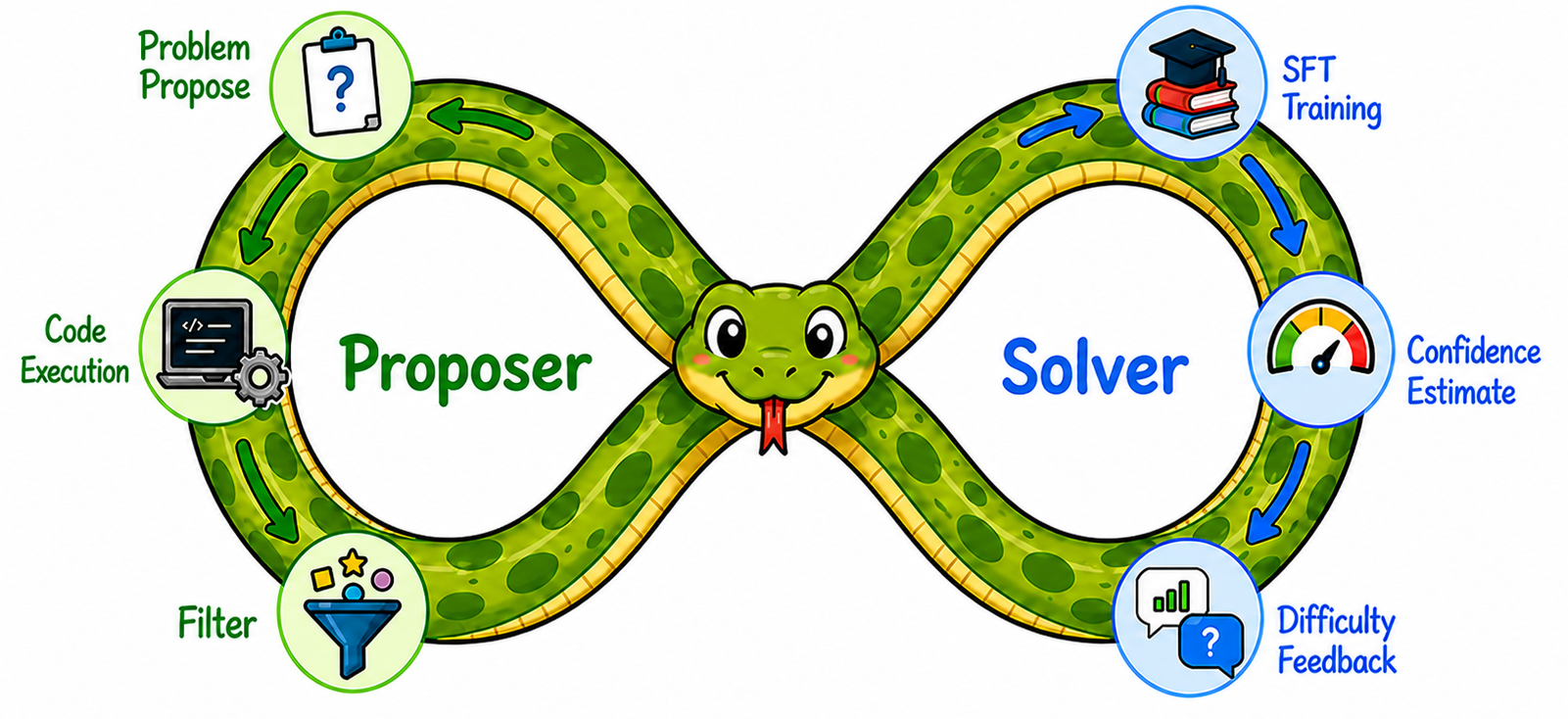}
    \caption{Overview of the \emph{Ouroboros-Spatial} framework. The proposer (left loop) generates spatial questions and programs, executes the programs to obtain answers, and filters the data, while the solver (right loop) learns from the curated data and provides difficulty feedback via confidence estimation.}
    \label{fig:ouro}
\end{figure}

In this work, we pursue a dynamic strategy for effective and efficient learning of spatial reasoning, taking a step toward more general spatial intelligence. Inspired by recent advances in self-evolving LLMs \cite{liang2025absolutezero,huang2025r}, we introduce \emph{Ouroboros-Spatial}, an iterative optimization framework in which a model alternates between two roles: a \emph{proposer} and a \emph{solver}. 
As illustrated in Figure~\ref{fig:ouro}, at each round, the proposer generates and filters spatial question--answer (QA) pairs, which are used to optimize the solver via supervised fine-tuning (SFT). The solver then estimates the difficulty of each QA pair based on its prediction confidence, feeding this signal back to the proposer to encourage new questions near the solver's evolving difficulty frontier. This closed loop allows the training distribution to adapt to the solver's current capability, enabling continuous improvement in spatial reasoning without additional human-curated data.

We apply the Ouroboros-Spatial pipeline to Qwen3-VL-4B and Qwen3-VL-8B. Using \emph{only 25.6k training samples}---10$\times$ to 100$\times$ fewer than recent curated datasets---our models achieve state-of-the-art average scores of $62.7$ and $63.3$ on VSI-Bench~\cite{yang2025thinking}, respectively. They outperform all open-source baselines within their size classes and surpass proprietary systems such as GPT-5~\cite{singh2025openaigpt5card} and Gemini-3-Pro~\cite{Gemini31}. Notably, our models also perform strongly on the debiased variant of VSI-Bench and show positive transfer on average across a diverse set of additional spatial reasoning benchmarks~\cite{yang2025mmsi,team2025gemini,wang2025mindcube,li2025viewspatial,du2024embspatial}, suggesting that the improvements arise from enhanced spatial reasoning rather than shortcut exploitation or overfitting to a specific evaluation distribution.

Our contributions are summarized as follows:
\begin{itemize}
\item %We propose \emph{Ouroboros-Spatial}, the first closed-loop, self-evolving framework for spatial reasoning in which a proposer generates difficulty-adaptive question--answer (QA) pairs and a solver learns from them, with the solver's confidence feedback continuously steering the proposer toward the model's learning frontier.
We propose \emph{Ouroboros-Spatial}, a closed-loop, self-evolving framework for spatial reasoning that, to our knowledge, is among the first to couple difficulty-adaptive QA generation with model optimization, using model confidence to steer generation toward the learning frontier.
\item We introduce a lightweight difficulty estimation mechanism based on the solver's token-level prediction confidence, enabling curriculum-aware data generation at no additional inference cost. Together with code-executed ground-truth derivation, the pipeline ensures both data quality and appropriate difficulty throughout training.
\item Extensive experiments show that Ouroboros-Spatial achieves state-of-the-art results on VSI-Bench with an order-of-magnitude less training data than prior work. Our models also demonstrate strong robustness on the debiased benchmark and positive transfer to other spatial reasoning benchmarks, validating the generality of the self-evolving paradigm for spatial intelligence.
\end{itemize}

%% file: chap/02-related_work.tex
\section{Related Work}
\label{sec:related}

\textbf{Multimodal Large Language Models and Spatial Reasoning.} Multimodal large language models (MLLMs) extend language models beyond text-only processing to understand and reason over multiple modalities, including text, images, videos, and audio. 
% Vision-language models (VLMs) represent a major branch of this paradigm. Early VLMs, such as LLaVA~\cite{li2025llava} and Qwen-VL~\cite{team2023qwen}, exploited a lightweight connector to align pretrained visual encoders with language models. 
Recently, multimodal reasoning has become a native capability of leading foundation models. Proprietary systems such as Gemini~\cite{Gemini31,team2025gemini}, GPT-5~\cite{singh2025openaigpt5card}, Seed-2.0~\cite{bytedance2025seed20}, and Kimi-K2.5~\cite{team2026kimi} have shown strong performance on challenging multimodal benchmarks~\cite{yue2025mmmu,wang2024charxiv,wang2024measuring}. 

However, despite this rapid progress, spatial reasoning remains a significant challenge and has become an active research area in multimodal learning. It requires MLLMs to understand complex spatiotemporal relationships, infer the geometric structure of objects and scenes, and reason about navigation in dynamic environments~\cite{liu2025spatial}. Recent advances in this field have been primarily driven by two major research directions. 

One direction focuses on spatially aware modeling and reasoning, where existing work can generally be grouped into three categories. The first category aims to recognize spatial and geometric relationships among objects from multiple images or videos through interaction with external visual tools or stronger teacher VLMs~\cite{wureinforcing,ma2026thinking,wu2026chatting,chen2025spacetools}. 
Rather than relying solely on 2D inputs, the second category goes one step further by reconstructing the underlying 3D structure of the scene from the 2D observations, often with the assistance of external 3D toolkits or foundation models~\cite{zhang2026think3d,chen2025think,gao2026map2thought,chen2026spatialcode,zhao2025spacemind,wu2025spatialmllm}. 
For instance,~\citet{zhang2026think3d} enabled a VLM to iteratively interact with 3D scenes through calls to a 3D Manipulation Toolkit.~\citet{chen2026spatialcode} transformed videos into explicit 3D spatial codes and fed them into a text-only LLM for downstream reasoning. 
Unlike the first two categories, which derive spatial information through passive visual processing, 
The third category adopts a more proactive strategy by simulating scenes beyond the static inputs and inferring spatial relationships from these auxiliary scenes with the help of visual generative models~\cite{li2025imagine,cao2025seeing,cao2025spatialdreamer,yangmindjourney}.
As representative examples,~\citet{yangmindjourney} employed a world model to simulate camera movements and generate ego-centric views as reasoning trajectories for image-question pairs. Similarly,~\citet{cao2025spatialdreamer} performed spatial reasoning by interleaving textual analysis with mental imagery rendered by an external world model.

In addition to methodological studies, the other major research direction advances spatial reasoning from the perspective of data. Indeed, the rapid progress of the field has benefited greatly from well-curated benchmark datasets, such as VSI-Bench~\cite{yang2025thinking}, MindCube~\cite{wang2025mindcube}, and MMSI-Bench~\cite{yang2025mmsi}. Following the emergence of benchmarks, the community has increasingly scaled up training data to improve model performance~\cite{ray2024sat,sensenova2024,feng2025vica,fan2025vlm3r,yang2024vst,yang2024cambrians,ouyang2025spacer,sun2025spacevista}. A common strategy is to leverage annotated video datasets, such as ScanNet~\cite{dai2017scannet}, ScanNet++~\cite{yeshwanth2023scannet++}, and ARKitScenes~\cite{baruch2021arkitscenes}, and then employ either handcrafted templates or large language models to synthesize question-answer pairs for instruction tuning~\cite{fan2025vlm3r,feng2025vica,yang2024cambrians,ouyang2025spacer}. Beyond annotated videos, recent work~\cite{yang2024cambrians} further incorporated simulated data and unlabeled videos to expand the scale and diversity of training data.

\textbf{Self-Evolving and Self-Play Training.} The training of large language models is undergoing a shift from human-supervised learning toward model self-evolution~\cite{fang2025comprehensive}. Early work such as STaR~\cite{zelikman2022star}, self-play training~\cite{chen2024spin}, and self-rewarding language models~\cite{yuan2024selfrewarding} showed that models can improve by learning from their own generated rationales, responses, or rewards, but still largely relied on human-crafted tasks for initialization. More recent studies go further by integrating task generation and model optimization into a unified iterative loop~\cite{liang2025absolutezero,huang2025r,chen2025self,kuba2025language,yang2025spell,liu2025spice,yue2026dr,wei2025toward,acikgoz2026tool}. A representative example is the proposer-solver paradigm of Absolute Zero~\cite{liang2025absolutezero}, where a model co-evolves as both task proposer and problem solver. This framework has since been extended to domains such as long-context modeling, search, software engineering, and tool calling~\cite{yang2025spell,yue2026dr,wei2025toward,acikgoz2026tool}.
More recently, advances in self-evolving language models have stimulated similar explorations in multi-modal models, leading to studies such as M-STaR~\cite{liu2025diving}, VisPlay~\cite{he2025visplay}, V-Zero~\cite{li2025vzero}, Vision-Zero~\cite{wang2025vision}, MM-Zero~\cite{zhao2025mmzero}, and EvoLMM~\cite{sun2025evolmm}. Most of these works follow the proposer-solver paradigm, but differ in how rewards are designed for the two roles.

\textbf{Relation to Existing Efforts.} In this work, we further advance the spatial reasoning capabilities of MLLMs through a data-centric perspective. Specifically, we adapt the proposer-solver self-evolving framework, originally developed for language reasoning, to spatial reasoning, enabling an effective yet highly efficient training paradigm. Unlike recent self-evolving VLMs that primarily focus on single-image scenarios, our framework is capable of generating high-quality spatial reasoning questions from videos, substantially expanding the complexity and diversity of training data. With this design, our method pushes state-of-the-art open-source VLMs to new performance frontiers across a wide range of benchmarks while requiring 10$\times$ to 100$\times$ less training data than recent large-scale data curation approaches. To the best of our knowledge, this is the first work to achieve such strong spatial reasoning gains under a self-evolving framework.

%% file: chap/03-method.tex
\section{Method}
\label{sec:method}
%\subsection{Overview}

Recent efforts to scale spatial intelligence~\cite{feng2025vica,fan2025vlm3r,yang2024cambrians} typically construct large-scale spatial question--answer (QA) datasets by applying rule-based programs to annotated 3D corpora~\cite{dai2017scannet,yeshwanth2023scannet++,baruch2021arkitscenes,yang2024vst}. While effective at scale, such static pipelines yield fixed corpora whose difficulty distributions are decoupled from the model being trained, inevitably mixing questions the model has already mastered with others far beyond its current capacity.

We propose \emph{Ouroboros-Spatial}, a fundamentally different approach to scaling spatial intelligence from a data-centric perspective. Specifically, it introduces a self-evolving pipeline that alternates between data generation and model fine-tuning, using the solver’s confidence on answer tokens as feedback to steer subsequent data generation toward the model’s difficulty frontier. Figure~\ref{fig:ouro} provides an overview. The pipeline proceeds iteratively, with each iteration comprising three stages: (1)~the proposer generates and filters spatial QA pairs (\S\ref{subsec:proposer}); (2)~the solver is fine-tuned on the accepted pairs and produces difficulty labels (\S\ref{subsec:solver}); and (3)~these labels are fed back to the proposer to guide the next round of data generation (\S\ref{subsec:feedback}). Algorithm~\ref{alg:spatialevo} summarizes the full Ouroboros-Spatial pipeline.

\begin{algorithm}[t]
\caption{Ouroboros-Spatial: Self-Evolving Training for Spatial Reasoning}
\label{alg:spatialevo}
\begin{algorithmic}[1]
\Require Scene pool $\mathcal{X} = \{s_1, \ldots, s_N\}$ with frames and 3D metadata;
        frozen proposer $\mathcal{P}$;
        pretrained solver $\mathcal{S}^{(0)}$;
        number of rounds $T$;
        difficulty thresholds $\tau_{\text{easy}}, \tau_{\text{hard}}$.
% \Statex \textbf{Notation:} $\mathcal{D}^{(t)}$ denotes the generated training dataset at round $t$.
\State $\text{feedback}^{(0)} \gets \varnothing$
\For{$t = 1$ to $T$}
  \State $\mathcal{D}^{(t)} \gets \varnothing$ \textcolor{gray}{\texttt{// generated training dataset at round $t$}}
  \State \textcolor{gray}{\texttt{// Stage 1: Propose and filter}}
  \For{each scene $s \in \mathcal{X}$}
    \State $\{(f(s_j), q_j, c_j)\} \gets \mathcal{P}\bigl(f(s), \text{meta}(s), \text{feedback}^{(t-1)}(s)\bigr)$, where $s_j = s$
    \Statex \hspace{\algorithmicindent}\hspace{\algorithmicindent}\hspace{\algorithmicindent}
            \textcolor{gray}{\texttt{// $q_j$: candidate question; $c_j$: answer-deriving program}}
    \For{each $j$}
      \State $a_j \gets \text{Execute}(c_j, \text{meta}(s_j))$; discard if execution fails
      \State $\text{accept/reject} \gets \mathcal{P}\bigl(f(s_j),\, q_j,\, a_j\bigr)$
      \State If accepted, $\mathcal{D}^{(t)} \gets \mathcal{D}^{(t)} \cup \{(f(s_j), q_j, a_j)\}$
    \EndFor
  \EndFor
  \State \textcolor{gray}{\texttt{// Stage 2: Train solver and estimate difficulty}}
  \State Fine-tune $\mathcal{S}^{(t-1)}$ on $\mathcal{D}^{(t)}$ for $K$ steps $\rightarrow \mathcal{S}^{(t)}$;
  \Statex \hspace{\algorithmicindent}\hspace{\algorithmicindent} record $p_j^{(t)}$ for each sample during training (Eq.~\eqref{eq:confidence})
  \For{each $(f(s_j), q_j, a_j) \in \mathcal{D}^{(t)}$}
    \State Assign $\text{difficulty}(j)$ via Eq.~\eqref{eq:difficulty} using recorded $p_j^{(t)}$
  \EndFor
  \State \textcolor{gray}{\texttt{// Stage 3: Compile feedback}}
  \State $\text{feedback}^{(t)}(s) \gets \text{Aggregate}(\{\text{difficulty}(j), q_j, a_j\}_{j: s_j = s})$ for each $s$
\EndFor
\State \Return Trained solver $\mathcal{S}^{(T)}$
\end{algorithmic}
\end{algorithm}

\begin{figure}[t]
\centering
\includegraphics[width=\textwidth]{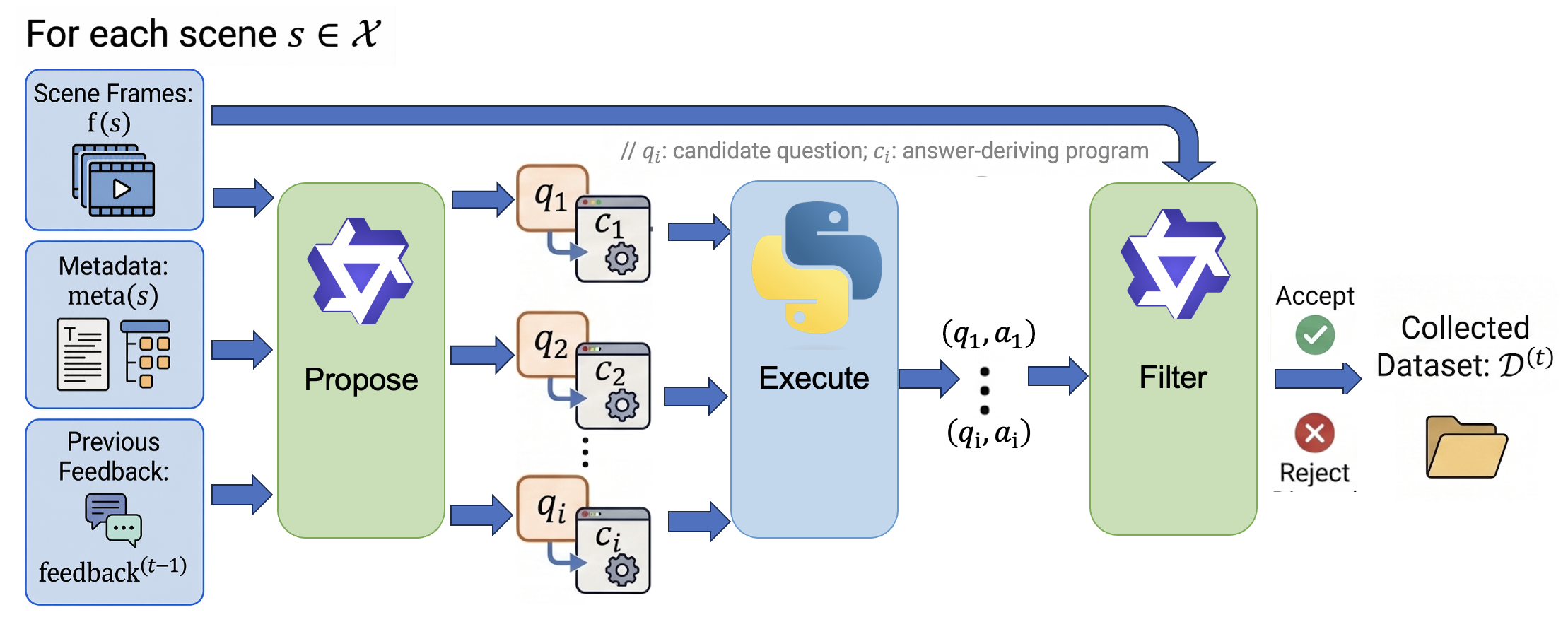}
\caption{The ``propose--execute--filter'' pipeline for question generation in Ouroboros-Spatial: the proposer first generates candidate question--program pairs; the programs are then executed to obtain ground-truth answers; finally, a filter verifies consistency with the visual frames before adding accepted samples to the training set $\mathcal{D}^{(t)}$.}%Stage~1 of the Ouroboros-Spatial pipeline: the proposer generates candidate question--program pairs, programs are executed to obtain ground-truth answers, and a filter verifies consistency against visual frames before accepting samples into the training set $\mathcal{D}^{(t)}$.}
\label{fig:proposer}
\end{figure}

\subsection{Proposer: Question Generation and Filtration}
\label{subsec:proposer}
\paragraph{Input representation.}
Following VLM-3R~\cite{fan2025vlm3r}, we construct a spatio-temporal scene graph for each indoor scene $s$ from the \emph{training set} of three open-source 3D datasets that provide 3D geometry, semantic, and instance meta-information~\cite{dai2017scannet, yeshwanth2023scannet++, baruch2021arkitscenes}. The scene graph consolidates per-frame object bounding boxes, semantic labels, 3D positions, and sizes into a unified metadata structure $\text{meta}(s)$. In parallel, we uniformly sample $N = 32$ frames $f(s)$ from the video of scene $s$ as visual context. In contrast to VLM-3R and other static pipelines that rely on hand-crafted templates to generate QA pairs from metadata alone, we feed both metadata and visual frames to an MLLM \emph{proposer}, which produces questions together with executable code for answer derivation.

\paragraph{QA generation.}
% In round $t$, for each scene $s$, a proposer $\mathcal{P}$ receives the frames, metadata, as well as the difficulty feedback from the previous round (for $t \geq 2$) and is tasked to produce a set of candidate questions $\{q_j\}$ associated with the corresponding \emph{answer-deriving program} $\{c_j\}$, where $s_j$ denotes the scene from which sample $j$ is drawn:
In round $t$, for each scene $s$, a proposer $\mathcal{P}$ takes as input the frames, metadata, and difficulty feedback from the previous round (for $t \geq 2$), and generates a set of candidate questions $\{q_j\}$ together with their corresponding \emph{answer-deriving programs} $\{c_j\}$:
\begin{equation}
  \mathcal{P}\bigl(f(s),\;\text{meta}(s),\;\text{feedback}^{(t-1)}(s)\bigr)
  \;\longrightarrow\;
  \bigl\{\,(f(s_j),\; q_j,\; c_j)\,\bigr\},\; \text{where } s_j = s.
  \label{eq:propose}
\end{equation}
The program is then executed against $\text{meta}(s_j)$ to obtain the ground-truth answer:
\begin{equation}
  a_j \;=\; \text{Execute}\bigl(c_j,\;\text{meta}(s_j)\bigr).
  \label{eq:execute}
\end{equation}
Since the answer is produced by deterministic code operating on structured
metadata rather than generated by an LLM, it is consistent with the available
metadata whenever the program executes successfully. The frames $f(s)$ and the VQA pairs are then sent to the proposer again to determine whether to accept each question. Figure~\ref{fig:proposer} illustrates the full propose--execute--filter pipeline. We find this filtering step to be essential: manual inspection of discarded questions confirms that the proposer correctly rejects problematic cases, including (1)~questions that are trivially solvable via language shortcuts without visual reasoning (e.g., ``How many bathtubs are there in the room? Answer:1''), and (2)~questions whose code-derived answers are incorrect due to metadata noise or mismatch, as exemplified in Appendix~\ref{appendix:case_study}.

\subsection{Solver: Fine-Tuning with Difficulty Estimation}
\label{subsec:solver}

\paragraph{Supervised fine-tuning.}
The solver $\mathcal{S}^{(t)}$ is initialized from the checkpoint of the
previous round (or from the base pretrained model when $t = 1$) and fine-tuned on
$\mathcal{D}^{(t)}$ for $K$ gradient steps with global batch size $B$.
The training objective is the standard next-token cross-entropy loss restricted
to the answer tokens; visual and question tokens participate in the forward pass
but do not contribute to the gradient.
%Let $a_j = (a_j^1, a_j^2, \ldots, a_j^{L_j})$ denote the $L_j$ ground-truth
%answer tokens for sample $j$. The loss is formulated as:
Let $a_j = (a_j^1, a_j^2, \ldots, a_j^{L_j})$ denote the ground-truth answer for sample $j$. The learning loss is formulated as:
\begin{equation}
  \mathcal{L}^{(t)} \;=\; -\frac{1}{|\mathcal{D}^{(t)}|}
  \sum_{(f(s_j),\,q_j,\,a_j)\in\mathcal{D}^{(t)}}
  \frac{1}{L_j}\sum_{\ell=1}^{L_j}
  \log P_{\mathcal{S}^{(t)}}\!\bigl(a_j^\ell \mid f(s_j),\, q_j,\, a_j^{<\ell}\bigr),
  \label{eq:sft_loss}
\end{equation}
where $a_j^\ell$ refers to the $\ell$-th token of $a_j$, $L_j$ is the number of tokens, and $s_j$ is the
scene from which the sample is drawn.

\paragraph{Difficulty estimation via prediction confidence.}
Because each answer is either a single option token (for multiple-choice
questions) or a short numeric string comprising only a few tokens
(e.g., ``140''), we can obtain a per-sample confidence score directly from the
forward pass already performed for Eq.~\eqref{eq:sft_loss}, requiring no
additional computation.
We define the confidence score as the geometric mean of the token-level
conditional probabilities produced during training:
\begin{equation}
  p_j^{(t)} \;=\;
  \exp\left(
  \frac{1}{L_j}\sum_{\ell=1}^{L_j}
  \log P_{\mathcal{S}^{(t)}}\!\bigl(a_j^\ell \mid f(s_j),\, q_j,\, a_j^{<\ell}\bigr)
  \right).
  \label{eq:confidence}
\end{equation}
After training completes, each sample is assigned a difficulty label based on its
recorded probability and two thresholds
$\tau_{\text{easy}}$ and $\tau_{\text{hard}}$:
\begin{equation}
  \text{difficulty}(j) \;=\;
  \begin{cases}
    \text{\textsc{easy}}  & \text{if } p_j^{(t)} > \tau_{\text{easy}}, \\
    \text{\textsc{hard}}  & \text{if } p_j^{(t)} < \tau_{\text{hard}}, \\
    \text{\textsc{frontier}} & \text{otherwise}.
  \end{cases}
  \label{eq:difficulty}
\end{equation}

\subsection{Feedback Compilation for Iterative Question Generation}
\label{subsec:feedback}
The final stage converts the per-sample difficulty labels from Stage~2 into
scene-specific feedback for the next round of question generation. For
each scene $s$, the feedback summary contains the previously generated questions,
their answers, and their difficulty labels:
\textsc{easy}, \textsc{hard}, or \textsc{frontier}. This scene-specific summary
is injected into the proposer's context in round $t{+}1$.

The purpose of this feedback is to reshape the next-round question distribution
according to the solver's current capability. In particular, the proposer is
instructed to reduce questions similar to \textsc{easy} samples, which the solver
has already mastered, as well as questions similar to \textsc{hard} samples,
which may be ambiguous, noisy, or beyond the solver's current capacity. This
encourages the proposer to generate more \textsc{frontier} questions and therefore provide more useful training signals for the evolving solver. The full feedback prompt template is
provided in Appendix~\ref{appendix:feedback_prompt}.

%% file: chap/04-experiment.tex
\section{Experiments}
\label{sec:experiment}
We evaluate Ouroboros-Spatial through extensive empirical studies.
\subsection{Experimental Setup}
\paragraph{Implementation Details.}
We apply the Ouroboros-Spatial pipeline (Algorithm~\ref{alg:spatialevo}) to two base models: Qwen3-VL-4B and Qwen3-VL-8B~\cite{bai2025qwen3vltechnicalreport}. In both settings, the proposer and solver are initialized from the same pretrained model. During training, only the solver is updated through supervised fine-tuning (SFT), while the proposer remains frozen throughout all rounds, with only its context evolving over iterations. For both models, we perform $T = 4$ iterative rounds. In each round, the solver is fine-tuned for $K = 100$ steps using a global batch size of $B = 64$, resulting in $25.6$k training samples in total. The difficulty thresholds are fixed at $\tau_{\text{easy}} = 0.9$ and $\tau_{\text{hard}} = 0.1$ across all rounds. We use a constant learning rate of $1 \times 10^{-6}$ without warm-up. To maintain optimization continuity, both the optimizer state and learning rate scheduler state are preserved across rounds as newly synthesized data is introduced. Additional training details and hyperparameter discussion are provided in Appendices~\ref{appendix:training_recipe} and~\ref{appendix:hyperparameter_discussion}, respectively.
\paragraph{Baselines.}
We compare our models against a broad and diverse set of baselines, including proprietary systems such as Gemini-3-Pro, Gemini-2.5-Pro~\cite{Gemini31}, GPT-5~\cite{singh2025openaigpt5card}, Seed-2.0~\cite{bytedance2025seed20}, and Grok-4~\cite{xai2025grok4}; open-source vision-language models (VLMs), including the Qwen3-VL series~\cite{bai2025qwen3vltechnicalreport}, InternVL3 series~\cite{zhu2025internvl3}, and LLaVA-OneVision~\cite{li2024llava}; as well as specialized open-source spatial intelligence models such as Cambrian-S~\cite{yang2024cambrians}, VST~\cite{yang2024vst}, ViCA~\cite{feng2025vica}, and Think with Spatial Code~\cite{chen2026spatialcode}. %Our primary evaluation is conducted on VSI-Bench and VSI-Debiased Bench~\cite{yang2025thinking}. To provide a more comprehensive assessment, we further evaluate all models on a diverse suite of additional spatial reasoning benchmarks, including ERQA~\cite{team2025gemini}, MindCube~\cite{wang2025mindcube}, MMSI~\cite{yang2025mmsi}, ViewSpatial~\cite{li2025viewspatial}, and EmbSpatial~\cite{du2024embspatial}.
\subsection{Main Results}
\paragraph{Improvements on Spatial Cognition.}
VSI-Bench~\cite{yang2025thinking} is constructed from the validation splits of ScanNet~\cite{dai2017scannet}, ScanNet++~\cite{yeshwanth2023scannet++}, and ARKitScenes~\cite{baruch2021arkitscenes}, comprising over 5,000 questions spanning eight spatial reasoning categories. It has become a widely adopted benchmark for comprehensive evaluation of multimodal large language models (MLLMs) on spatial relationship understanding, metric estimation, and higher-order spatial reasoning. Following the original evaluation protocol~\cite{yang2025thinking}, we report Mean Relative Accuracy (MRA) for numerical questions and Accuracy (ACC) for multiple-choice questions, with the final overall score computed as the macro-average across all categories. Consistent with previous work~\cite{bai2025qwen3vltechnicalreport}, we uniformly sample 32 frames from each scene during evaluation.

Table~\ref{tab:main_results} reports results on VSI-Bench. In terms of the overall performance, Ouro-Spatial-4B achieves the best average score of 62.7 among 3B--4B spatial models. Scaling to 8B, Ouro-Spatial-8B further improves to 63.3, establishing a new state of the art. Notably, both models are trained with \emph{only 25.6k} samples, corresponding to a 10$\times$ to 100$\times$ improvement in data efficiency over prior data curation efforts~\citep{feng2025vica,yang2024cambrians,fan2025vlm3r}.

At the category level, the gains are particularly pronounced on Room Size, Object Size, and Object Count. Although these categories are relatively easy to instantiate using templates, exhaustive template-based generation often yields a large number of trivial questions (e.g., typical room dimensions, common object sizes, or counts of a few salient objects). Such questions can be answered using dataset-level regularities or everyday priors, without requiring precise visual grounding, and may therefore reinforce shortcut behaviors rather than improve genuine spatial reasoning. In contrast, Ouro-Spatial leverages solver feedback to adaptively reshape the training distribution as the model evolves. Questions that become too easy are down-weighted in subsequent rounds, while generation is steered toward samples near the solver’s current difficulty frontier. This curriculum-like adaptation produces more informative supervision, enabling stronger performance with substantially fewer training examples.

Note that our training data has \emph{no overlap} with the evaluation benchmark: although both are derived from the same underlying 3D datasets, we exclusively use the training splits, while the benchmark is constructed from the validation splits. We further note that the ``Route Planning'' category is excluded from training, as non-trivial instances cannot be reliably generated and verified from scene-graph metadata alone and would require costly annotations. Consequently, performance on this category remains modest.

\definecolor{smallbg}{RGB}{232,243,255}
\definecolor{largebg}{RGB}{255,239,224}

\begin{table}[t]
\centering
\caption{Results on VSI-Bench. Best results in each size group are highlighted in \textbf{bold}.
$\ddagger$: uses 2D bounding box annotations as additional input, which helps a lot for tasks like object counting and relative distance; without them the Avg.\ drops to 57.0.
$\dagger$: originally trained and evaluated on $128$ frames; we re-evaluate on $32$ frames for fair comparison.} %Rows are tinted \colorbox{smallbg}{\strut\,blue\,} for 3B--4B  models and
%\colorbox{largebg}{\strut\,orange\,} for 7B--8B models.}
\label{tab:main_results}
\resizebox{\textwidth}{!}{%
\begin{tabular}{l c c c c c c c c c}
\toprule
\textbf{Model}
  & \textbf{Avg.}
  & \textbf{Obj.Cnt}
  & \textbf{Abs.Dist}
  & \textbf{Obj.Size}
  & \textbf{RoomSz}
  & \textbf{Rel.Dis}
  & \textbf{Rel.Dir}
  & \textbf{Route}
  & \textbf{App.Ord} \\
\midrule
\multicolumn{10}{l}{\textit{Reference}} \\
Human
  & 79.2 & 94.3 & 47.0 & 60.4 & 45.9 & 94.7 & 95.8 & 95.8 & 100.0 \\
Random
  & 34.0 & 62.1 & 32.0 & 29.9 & 33.1 & 25.1 & 47.9 & 28.4 & 25.2 \\
\midrule
\multicolumn{10}{l}{\textit{Proprietary Models}} \\
Seed-2.0
  & 50.7 & 49.4 & 25.3 & 69.5 & 25.8 & 61.8 & 44.9 & 44.3 & 71.0 \\
Grok-4
  & 47.9 & 37.1 & 32.9 & 60.8 & 48.4 & 53.1 & 39.6 & 47.3 & 66.8 \\
Gemini-2.5-Pro
  & 53.5 & 46.0 & 37.3 & 68.7 & 54.3 & 61.9 & 43.9 & 47.4 & 68.7 \\
Gemini-3-Pro
  & 56.0 & 49.0 & 42.8 & 71.5 & 41.8 & 56.6 & 57.5 & 61.9 & 68.0 \\
Kimi-K2.5
  & 53.6 & 57.2 & 34.9 & 69.3 & 54.4 & 59.6 & 41.3 & 52.1 & 67.0 \\
GPT-5
  & 55.0 & 53.3 & 34.4 & 73.3 & 48.3 & 47.8 & 48.6 & 50.2 & 68.9 \\
\midrule
\multicolumn{10}{l}{\textit{Open-Source General Models}} \\
InternVL3-2B
  & 32.9 & 64.8 & 30.8 & 32.4 & 22.9 & 32.2 & 34.3 & 32.9 & 12.6 \\
InternVL3-8B
  & 41.2 & 66.0 & 30.4 & 48.3 & 43.6 & 48.2 & 40.8 & 39.3 & 26.2 \\
LLaVA-OneVision-7B
  & 32.4 & 47.7 & 20.2 & 47.4 & 12.3 & 42.5 & 35.2 & 29.4 & 24.4 \\
Qwen2.5-VL-3B
  & 29.0 & 24.3 & 24.7 & 31.7 & 22.6 & 38.3 & 42.6 & 26.3 & 21.2 \\
Qwen2.5-VL-7B
  & 31.4 & 40.9 & 14.8 & 43.4 & 47.0 & 35.8 & 40.1 & 33.0 & 29.8 \\
Qwen3-VL-4B
  & 52.8 & 61.5 & 45.0 & 73.8 & 45.5 & 52.4 & 46.7 & 34.0 & 63.4 \\
Qwen3-VL-8B
  & 56.5 & 66.0 & 47.1 & 75.9 & 57.2 & 58.3 & 51.5 & 31.4 & 64.2 \\
\midrule
\rowcolor{smallbg}
\multicolumn{10}{l}{\textit{Open-Source Spatial Intelligence Models (3B-4B)}} \\
\rowcolor{smallbg}
MindCube-3B
  & 17.2 & 12.8 & 22.7 & 4.3 & 23.4 & 20.2 & 15.7 & 15.9 & 22.4 \\
\rowcolor{smallbg}
SpatiaLadder-3B
  & 44.8 & 62.1 & 35.3 & 61.9 & 41.4 & 45.6 & 46.4 & 27.3 & 38.5 \\
\rowcolor{smallbg}
VST-3B-SFT
  & 57.9 & 69.3 & 45.4 & 71.8 & 62.4 & 59.0 & 46.0 & \textbf{38.7} & 70.2 \\
\rowcolor{smallbg}
Cambrian-S-3B$^{\dagger}$
  & 57.3 & 70.7 & 40.6 & 68.0 & 46.3 & \textbf{64.8} & 61.9 & 27.3 & \textbf{78.8} \\
\rowcolor{smallbg}
Spatial-MLLM-4B
  & 47.0 & 65.3 & 34.8 & 63.1 & 45.1 & 41.3 & 46.9 & 33.5 & 46.3 \\
\rowcolor{gray!15}
\textcolor{gray}{Spatial Code-4B$^{\ddagger}$}
  & \textcolor{gray}{60.0} & \textcolor{gray}{92.0} & \textcolor{gray}{60.7} & \textcolor{gray}{50.8} & \textcolor{gray}{33.1} & \textcolor{gray}{62.0} & \textcolor{gray}{87.1} & \textcolor{gray}{32.5} & \textcolor{gray}{59.0} \\
\rowcolor{smallbg}
\textbf{Ouro-Spatial-4B}
  & \textbf{62.7} & \textbf{71.5} & \textbf{47.5} & \textbf{77.1} & \textbf{73.5} & 63.8 & \textbf{63.7} & 33.0 & 71.4 \\
\midrule
\rowcolor{largebg}
\multicolumn{10}{l}{\textit{Open-Source Spatial Intelligence Models (7B--8B)}} \\
\rowcolor{largebg}
SpaceR-7B
  & 41.5 & 44.5 & 24.7 & 53.5 & 37.6 & 31.9 & 46.1 & 29.3 & 54.8 \\
\rowcolor{largebg}
ViLaSR-7B
  & 44.6 & 58.1 & 33.3 & 61.4 & 28.8 & 48.5 & 46.5 & 29.9 & 53.2 \\
\rowcolor{largebg}
VST-7B-SFT
  & 60.6 & \textbf{72.0} & 44.4 & 74.3 & 68.3 & 59.7 & 55.8 & \textbf{44.9} & 65.2 \\
\rowcolor{largebg}
VLM-3R-7B
  & 60.9 & 70.2 & \textbf{49.4} & 69.2 & 67.1 & 65.4 & \textbf{80.5} & 45.4 & 40.1 \\
\rowcolor{largebg}
Cambrian-S-7B$^{\dagger}$
  & 62.9 & 68.2 & 45.8 & 72.5 & 67.6 & \textbf{66.8} & 69.6 & 39.2 & \textbf{73.8} \\
\rowcolor{largebg}
\textbf{Ouro-Spatial-8B}
  & \textbf{63.3} & 71.7 & 48.7 & \textbf{77.3} & \textbf{75.7} & 64.2 & 62.6 & 35.6 & 70.7 \\
\bottomrule
\end{tabular}}
\end{table}

\paragraph{Robustness on VSI-debiased.}
To verify that our models acquire genuine spatial understanding rather than exploiting language priors (e.g., memorizing that a typical desk is roughly 1.5\,m wide), we evaluate on VSI-debiased~\cite{yang2025thinking}, a variant of VSI-Bench specifically designed to eliminate such language shortcuts. Ouro-Spatial-4B and Ouro-Spatial-8B score 56.4 and 57.0 on the debiased split, dropping by only 6.3 points from their original VSI-Bench scores of 62.7 and 63.3, respectively. Notably, these debiased scores still surpass the best proprietary models on the original VSI-Bench, further confirming that the improvements from our iterative pipeline reflect robust spatial cognition rather than shortcut memorization.

\paragraph{Results on More Spatial Benchmarks.}
To evaluate whether Ouroboros-Spatial generalizes beyond VSI-Bench, we assess
Ouro-Spatial models on five additional benchmarks:
MindCube~\cite{wang2025mindcube}, ERQA~\cite{team2025gemini},
MMSI~\cite{yang2025mmsi}, ViewSpatial~\cite{li2025viewspatial}, and
EmbSpatial~\cite{du2024embspatial}.
Although these benchmarks are broadly categorized as ``spatial reasoning,''
they differ substantially from VSI-Bench in task format, visual input, and the
spatial skills they emphasize. For instance, MindCube tests mental rotation
with synthetic cube images, ERQA requires egocentric room-level question
answering, and MMSI contains a large proportion of camera-centric spatial
problems.
Crucially, \emph{our generation objectives do not explicitly target the
specific task formats or evaluation protocols of these benchmarks.}
Table~\ref{tab:other_benchmarks} reports accuracy for both the base models and
Ouro-Spatial variants. Despite the domain gap, both variants achieve positive
average gains and improve on four out of five benchmarks. These results suggest
that the iterative self-evolution pipeline strengthens general spatial cognition
rather than overfitting to a single benchmark.

\begin{table}[t]
\centering
\caption{Accuracy on other spatial benchmarks.
Changes from the corresponding base models are shown in
{\scriptsize\textcolor{green!50!black}{green}} /
{\scriptsize\textcolor{red!70!black}{red}}.}
\label{tab:other_benchmarks}
\begin{tabular}{l c c c c c c}
\toprule
\textbf{Model} & \textbf{Avg.} & \textbf{MindCube} & \textbf{ERQA}
& \textbf{MMSI} & \textbf{ViewSpatial} & \textbf{EmbSpatial} \\
\midrule
Qwen3-VL-4B
& 42.8
& 28.4
& 41.2
& 28.3
& 39.4
& 76.9 \\

Ouro-Spatial-4B
& 44.3\,{\scriptsize\textcolor{green!50!black}{+1.5}}
& 33.4\,{\scriptsize\textcolor{green!50!black}{+5.0}}
& 40.5\,{\scriptsize\textcolor{red!70!black}{$-$0.7}}
& 29.7\,{\scriptsize\textcolor{green!50!black}{+1.4}}
& 40.8\,{\scriptsize\textcolor{green!50!black}{+1.4}}
& 77.3\,{\scriptsize\textcolor{green!50!black}{+0.4}} \\
\midrule
Qwen3-VL-8B
& 45.7
& 35.0
& 43.0
& 31.1
& 41.6
& 77.6 \\

Ouro-Spatial-8B
& 46.9\,{\scriptsize\textcolor{green!50!black}{+1.3}}
& 35.4\,{\scriptsize\textcolor{green!50!black}{+0.4}}
& 44.0\,{\scriptsize\textcolor{green!50!black}{+1.0}}
& 30.5\,{\scriptsize\textcolor{red!70!black}{$-$0.6}}
& 46.0\,{\scriptsize\textcolor{green!50!black}{+4.4}}
& 78.8\,{\scriptsize\textcolor{green!50!black}{+1.2}} \\
\bottomrule
\end{tabular}
\end{table}

\subsection{Discussions}
Beyond the extensive evaluation across diverse benchmarks, we further analyze Ouroboros-Spatial from the following perspectives: (1) how performance evolves over iterations; (2) the contribution of individual components in the framework; and (3) performance on general video and multi-image understanding tasks.
%\subsubsection{Ablation Study}

\subsubsection{Performance over Iterations}

\begin{wraptable}{r}{0.42\textwidth}
\vspace{-1.2em}
\centering
\caption{VSI-Bench score across rounds.}
\label{tab:iterative_progress}
\vspace{0.6em}
\setlength{\tabcolsep}{6pt}
\begin{tabular}{l c c}
\toprule
\textbf{Round} & \textbf{Ouro-4B} & \textbf{Ouro-8B} \\
\midrule
Base    & 52.8 & 56.5 \\
Round 1 & 56.4 & 58.8 \\
Round 2 & 59.3 & 60.8 \\
Round 3 & 61.4 & 61.6 \\
Round 4 & \textbf{62.7} & \textbf{63.3} \\
\bottomrule
\end{tabular}
\vspace{-1em}
\end{wraptable}

Table~\ref{tab:iterative_progress} tracks the overall performance on VSI-Bench across iteration rounds. Both Ouro-Spatial-4B and Ouro-Spatial-8B improve steadily over four rounds. %achieving gains of $+9.9$ and $+6.8$ points over their respective baselines. 
The improvements are larger in early rounds and gradually taper, consistent with the pipeline progressively exhausting easy gains. Extending to a fifth round yields negligible improvement (Ouro-Spatial-8B: 63.36 at round~5 vs.\ 63.31 at round~4), suggesting that, under the current metadata and training pipeline, further optimization provides limited benefit.

\subsubsection{Ablation Study}
We study two key components of Ouroboros-Spatial. First, we evaluate the benefit of our LLM-based propose--execute--filter pipeline by comparing against ViCA-322k~\citep{feng2025vica}, a static template-generated spatial instruction-tuning corpus built from the same annotated 3D data sources used in our work, making it a natural comparison for isolating the effect of data-generation strategy.
Second, we isolate the role of difficulty feedback by removing the solver's difficulty labels from the next-round prompt. Note that we still provide the model with previously generated questions for the same scene to reduce duplicate generation. More details about the comparison of the data-source and the question-type distributions between Ouro-Spatial and ViCA-322k are provided in Appendix~\ref{appendix:data_composition}.

Table~\ref{tab:data_ablation} reports the results. Under compute-matched settings, Ouro-Spatial outperforms ViCA by a large margin, showing that LLM-based question generation with execution and visual filtering is more sample-efficient than static templates. Even when ViCA is trained on the full 322k corpus, our method remains ahead by 2.3 points on average. Removing difficulty feedback reduces performance across most categories. This further supports that the propose--execute--filter pipeline yields high-quality supervision, and that difficulty feedback provides an additional gain by adapting the generated data to the evolving solver.

\begin{table}[t]
\centering
\caption{Results for ablation study. All models are implemented with Qwen3-VL-4B as the base.
\emph{Compute-matched}: same number of samples as Ouro-Spatial.
\emph{Full}: trained on the entire ViCA-322k dataset.}
\label{tab:data_ablation}
\resizebox{\textwidth}{!}{%
\begin{tabular}{l r r c c c c c c c c c}
\toprule
\textbf{Training Data}
  & \textbf{\#Samples}
  & \textbf{Steps}
  & \textbf{Avg.}
  & \textbf{Obj.Cnt}
  & \textbf{Abs.Dist}
  & \textbf{ObjSz}
  & \textbf{RoomSz}
  & \textbf{Rel.Dist}
  & \textbf{Rel.Dir}
  & \textbf{Route}
  & \textbf{App.Ord} \\
\midrule
None (base model)
  & --   & --   & 52.8 & 61.5 & 45.0 & 73.8 & 45.5 & 52.4 & 46.7 & 34.0 & 63.4 \\
ViCA-322k (compute-matched)
  & 25.6k & 400  & 57.4 & 69.8 & 45.6 & 73.9 & 59.5 & 59.6 & 44.6 & 34.5 & 71.7 \\
ViCA-322k (full)
  & 322k & $\sim$5k & 60.4 & 72.7 & 53.0 & 77.7 & 74.3 & 58.3 & 46.4 & 27.8 & 73.0 \\
Ouro-Spatial (w/o difficulty feedback) 
  & 25.6k & 400 & 61.2 & 69.3 & 45.5 & 77.6 & 68.5 & 62.4 & 60.8 & 35.5 & 70.2 \\
\textbf{Ouro-Spatial (ours)}
  & 25.6k & 400 & 62.7 & 71.5 & 47.5 & 77.1 & 73.5 & 63.8 & 63.7 & 33.0 & 71.4 \\
\bottomrule
\end{tabular}}
\end{table}

\subsubsection{Performance on General Video and Multi-Image Benchmarks}
\label{subsec:general_bench}

One may be concerned that the performance gains on spatial benchmarks come at the expense of general video and multi-image understanding capabilities. To investigate this, we further evaluate Ouro-Spatial on four additional benchmarks: VideoMME~\cite{fu2025video} and MVBench~\cite{li2024mvbench} for video understanding, and MUIRBench~\cite{wang2024muirbench} and BLINK~\cite{fu2024blink} for multi-image reasoning. All video benchmarks are evaluated using 32 uniformly sampled frames, the same setting as VSI-Bench. As shown in Table~\ref{tab:general_bench}, Ouro-Spatial-4B and Ouro-Spatial-8B maintain, and in several cases slightly improve, performance relative to their base models on these general-purpose benchmarks. Overall, the average score across all four benchmarks remains comparable, confirming that Ouroboros-Spatial's self-evolving training strengthens spatial cognition without compromising the model's broader multimodal capabilities.

\begin{table}[h]
\centering
\caption{Performance on general video and multi-image benchmarks}
\label{tab:general_bench}
\resizebox{0.85\textwidth}{!}{%
\begin{tabular}{l *{5}{>{\centering\arraybackslash}p{1.8cm}}}
\toprule
\textbf{Model} & \textbf{VideoMME} & \textbf{MVBench} & \textbf{MUIRBench} & \textbf{BLINK} & \textbf{Avg.} \\
\midrule
Qwen3-VL-4B          & 62.9 & 65.8 & 59.0 & 64.3 & 63.0 \\
Ouro-Spatial-4B      & 63.9 & 66.4 & 58.2 & 63.1 & 62.9 \\
\midrule
Qwen3-VL-8B          & 65.5 & 66.5 & 60.0 & 64.5 & 64.1 \\
Ouro-Spatial-8B      & 66.5 & 67.5 & 58.9 & 64.2 & 64.3 \\
\bottomrule
\end{tabular}}
\end{table}

%% file: chap/05-conclusion.tex
\section{Conclusion}
\label{sec:conclusion}
In this work, we introduce Ouroboros-Spatial: a novel framework that closes the data-model loop to enhance the spatial reasoning ability for multimodal large language models (MLLMs). Using only 25.6k samples, our Ouro-Spatial models achieve state-of-the-art performance on VSI-Bench, surpassing models trained on 10--100$\times$ more data. The models further demonstrate robustness on the debiased benchmark and positive transfer to five additional spatial reasoning evaluations. We hope Ouroboros-Spatial offers a practical and data-efficient recipe for advancing spatial intelligence in MLLMs.

%% file: chap/06-appendix.tex
\section{Implementation Details}
\label{appendix:impl}
\subsection{Difficulty Feedback Prompt}
\label{appendix:feedback_prompt}
Starting from round $t \geq 2$, the proposer's prompt is augmented with
scene-specific feedback derived from the previous round. For each scene, we list
the questions that have already been generated, together with their answers, to
discourage duplicate question generation. In addition, each question is annotated
with its difficulty label. Questions labeled as \textsc{easy} indicate patterns
that the solver has already mastered, while questions labeled as \textsc{hard}
may be ambiguous, noisy, or beyond the solver's current capability. The proposer
is instructed to avoid generating questions that are too similar to these
difficulty extremes and to focus on more informative frontier-style questions.
The template is shown below.

\begin{quote}
\ttfamily\small
\#\# Previously Generated Questions for This Scene\\[4pt]
-- Question: \{question 1\}\\
\phantom{-- }Answer: \{answer 1\}\\
\phantom{-- }Difficulty: \{\textsc{easy}/\textsc{frontier}/\textsc{hard}\}\\
-- Question: \{question 2\}\\
\phantom{-- }Answer: \{answer 2\}\\
\phantom{-- }Difficulty: \{\textsc{easy}/\textsc{frontier}/\textsc{hard}\}\\
-- ...\\[6pt]
\#\# Difficulty Guidance\\[4pt]
Avoid generating questions that are too similar to \textsc{easy} questions, since the model has already mastered them.\\
Avoid generating questions that are too similar to \textsc{hard} questions, since they may be ambiguous, noisy, or beyond the model's current capability.\\
Focus on generating informative questions near the model's current frontier.
\end{quote}

\subsection{Training Recipe}
\label{appendix:training_recipe}
We fine-tune the solver using MS-Swift~\cite{zhao2024swiftascalablelightweightinfrastructure}\footnote{\url{https://github.com/modelscope/ms-swift/}} with full-parameter updates (no adapters or frozen layers). All experiments are conducted on 8 H200 GPUs. Table~\ref{tab:training_recipe} summarizes the key hyperparameters.

\begin{table}[h]
\centering
\caption{Solver fine-tuning hyperparameters.}
\label{tab:training_recipe}
\begin{tabular}{ll}
\toprule
\textbf{Hyperparameter} & \textbf{Value} \\
\midrule
Training framework       & MS-Swift \\
Parameter update         & Full fine-tuning \\
Number of GPUs           & 8 \\
Per-GPU batch size       & 8 \\
Global batch size        & 64 \\
Learning rate            & $1 \times 10^{-6}$ \\
LR scheduler             & Constant \\
Maximum sequence length  & 16\,384 \\
Distributed strategy     & DeepSpeed ZeRO-2 \\
\bottomrule
\end{tabular}
\end{table}

\subsection{Hyperparameter Discussion}
\label{appendix:hyperparameter_discussion}
We conduct two lightweight checks on Ouro-Spatial-8B: extending training to a fifth round yields 63.36 on VSI-Bench, nearly unchanged from 63.31 at round~4, while changing the difficulty thresholds to $\tau_{\text{hard}}=0.2$ and $\tau_{\text{easy}}=0.8$ yields 62.85, suggesting that performance largely saturates after four rounds and remains reasonably robust to threshold choice.

\subsection{Evaluation Prompt}
\label{appendix:eval_prompt}
Following Qwen3-VL tech report~\cite{Qwen35blog}, we use the following prompt templates for VSI-Bench evaluation.

\paragraph{Multiple-choice.}
\begin{quote}
\ttfamily\small
<video> These are frames of a video. \{question\} Options: \{options\} Answer with the option's letter from the given choices directly.
\end{quote}

\paragraph{Open-ended.}
\begin{quote}
\ttfamily\small
<video> These are frames of a video. \{question\} Please answer the question using a single word or phrase.
\end{quote}

\section{Additional Experiments}
\label{appendix:experiments}

\subsection{Training Data Composition}
\label{appendix:data_composition}
\begin{figure}[h]
  \centering
  \includegraphics[width=0.95\linewidth]{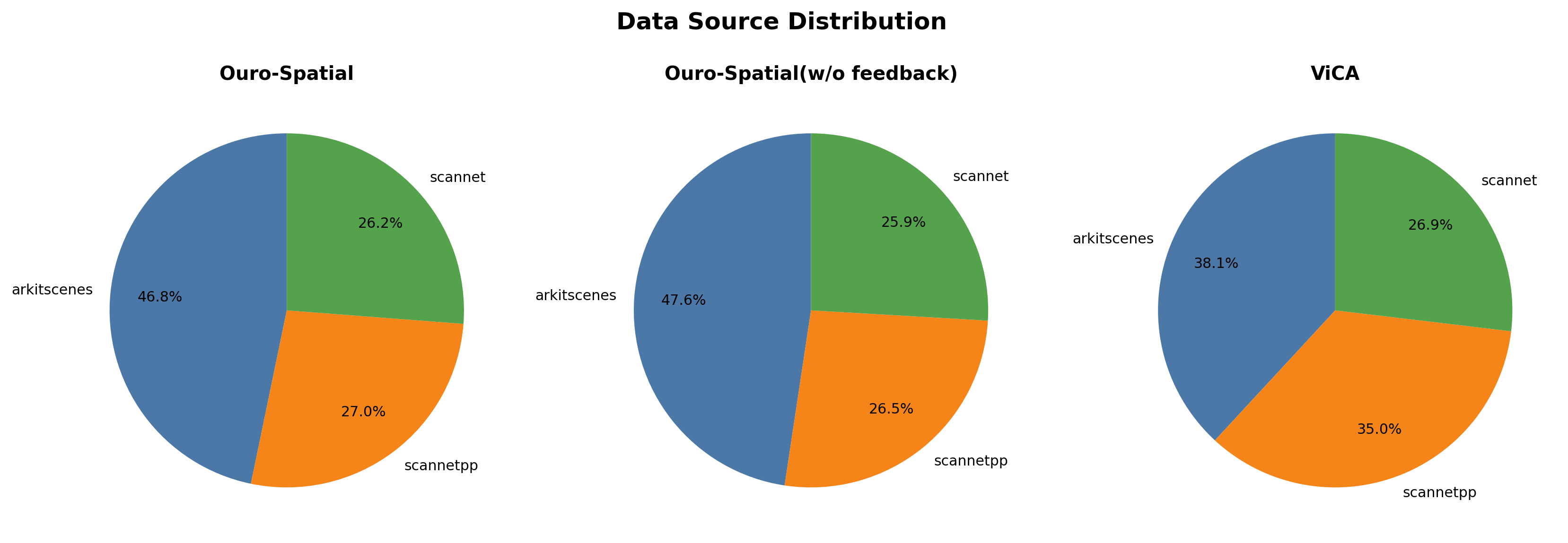}
  \caption{Comparison of data-source composition between Ouro-Spatial and ViCA-322k.}
  \label{fig:data_source_composition}
\end{figure}

Figure~\ref{fig:data_source_composition} compares the source distribution of Ouro-Spatial and ViCA-322k~\cite{feng2025vica}. Both datasets are built from indoor 3D video sources, but they differ substantially in scale and construction. Ouro-Spatial includes a larger proportion of questions from ARKitScenes, primarily because ARKitScenes contains more scenes than the other source datasets. 

\begin{figure}[h]
  \centering
  \includegraphics[width=0.95\linewidth]{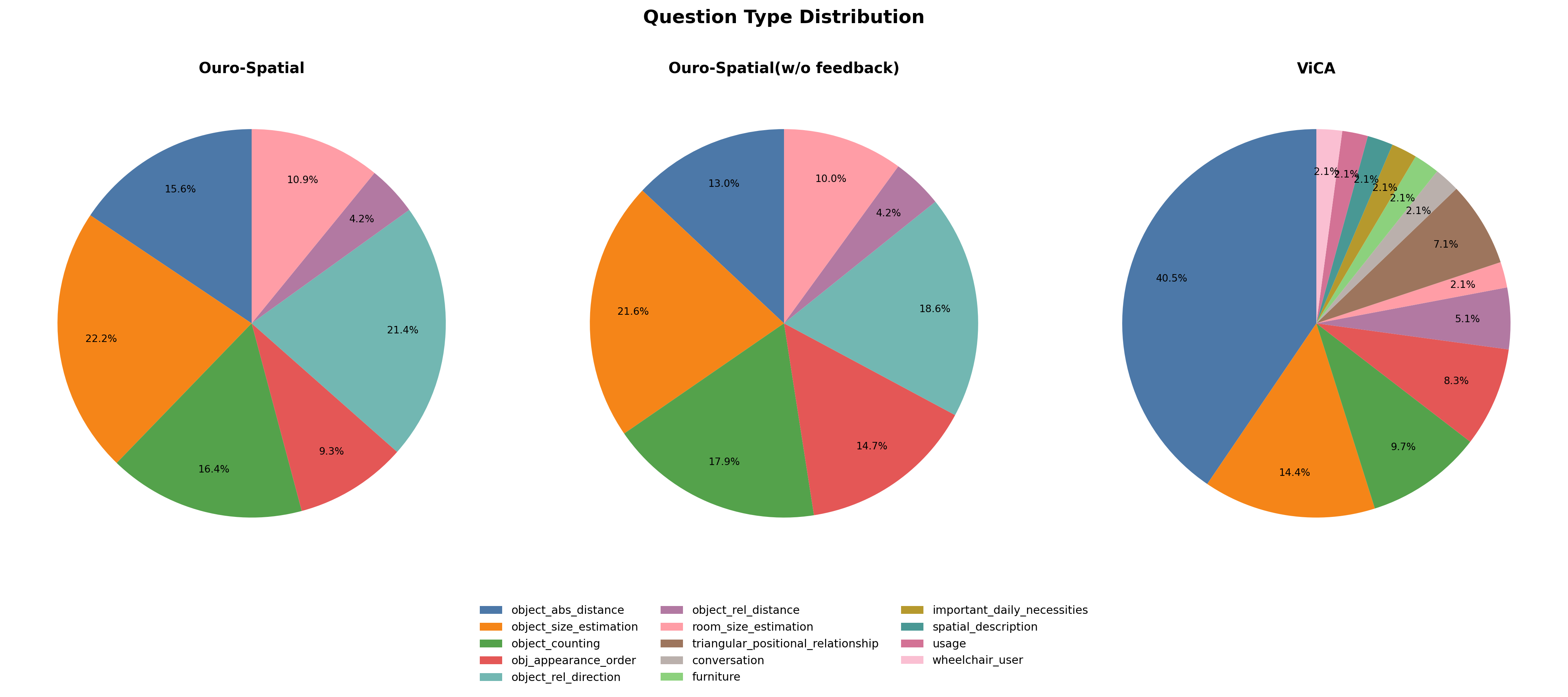}
  \caption{Normalized question-type composition of Ouro-Spatial and ViCA-322k. ViCA includes both metadata-grounded base tasks and additional language-grounded complex spatial cognition tasks beyond the VSI-style categories.}
  \label{fig:question_type_composition}
\end{figure}

Figure~\ref{fig:question_type_composition} further compares the normalized question-type distributions. Besides the VSI-style spatial categories, ViCA-322k contains additional question families described on its dataset card\footnote{\url{https://huggingface.co/datasets/nkkbr/ViCA-322K}}. Its base split includes six metadata-grounded spatial cognition tasks: object count, object relative distance, object size estimation, object absolute distance, object appearance order, and room size. For ARKitScenes, ViCA additionally provides a triangular positional relationship split, where each question asks for the side lengths and angles of the triangle formed by three specified objects. ViCA also includes a complex spatial cognition subset with open-ended, language-grounded tasks, including multi-turn spatial conversations, furniture-oriented questions, daily-necessity reasoning, spatial descriptions, usage-oriented questions, and wheelchair-user accessibility questions. 

\subsection{Per-Round Results on VSI-Bench}
\label{appendix:per_round}
To illustrate how the solver improves across self-evolution rounds, we report per-task VSI-Bench results for each round of training.

\begin{table}[h]
\centering
\caption{Per-task VSI-Bench results across rounds for \textbf{Ouro-Spatial-4B}.}
\label{tab:per_round_4b}
\resizebox{\textwidth}{!}{%
\begin{tabular}{l c c c c c c c c c}
\toprule
\textbf{Round} & \textbf{Avg.} & \textbf{Obj.Cnt} & \textbf{Abs.Dist} & \textbf{Obj.Sz} & \textbf{RoomSz} & \textbf{Rel.Dist} & \textbf{Rel.Dir} & \textbf{Route} & \textbf{App.Ord} \\
\midrule
Base   & 52.8 & 61.5 & 45.0 & 73.8 & 45.5 & 52.4 & 46.6 & 34.0 & 63.4 \\
Round 1 & 56.4 & 53.9 & 43.5 & 77.4 & 61.8 & 60.7 & 55.6 & 28.9 & 69.1 \\
Round 2 & 59.3 & 59.4 & 44.4 & 76.8 & 72.2 & 62.0 & 59.4 & 29.9 & 70.2 \\
Round 3 & 61.4 & 70.2 & 46.5 & 77.0 & 70.8 & 63.5 & 62.0 & 30.4 & 70.4 \\
Round 4 & 62.7 & 71.5 & 47.5 & 77.1 & 73.5 & 63.8 & 63.7 & 33.0 & 71.4 \\
\bottomrule
\end{tabular}}
\end{table}

\begin{table}[h]
\centering
\caption{Per-task VSI-Bench results across rounds for \textbf{Ouro-Spatial-8B}.}
\label{tab:per_round_8b}
\resizebox{\textwidth}{!}{%
\begin{tabular}{l c c c c c c c c c}
\toprule
\textbf{Round} & \textbf{Avg.} & \textbf{Obj.Cnt} & \textbf{Abs.Dist} & \textbf{Obj.Sz} & \textbf{RoomSz} & \textbf{Rel.Dist} & \textbf{Rel.Dir} & \textbf{Route} & \textbf{App.Ord} \\
\midrule
Base   & 56.5 & 66.0 & 47.1 & 75.9 & 57.2 & 58.3 & 51.5 & 31.4 & 64.2 \\
Round 1 & 58.8 & 70.0 & 45.2 & 76.1 & 67.1 & 61.3 & 52.1 & 29.4 & 69.4 \\
Round 2 & 60.8 & 68.9 & 49.5 & 77.3 & 69.0 & 62.3 & 56.7 & 32.5 & 70.7 \\
Round 3 & 61.6 & 71.0 & 49.3 & 77.0 & 69.6 & 62.8 & 60.9 & 32.5 & 70.1 \\
Round 4 & 63.3 & 71.7 & 48.7 & 77.3 & 75.7 & 64.2 & 62.6 & 35.6 & 70.7 \\
\bottomrule
\end{tabular}}
\end{table}

\subsection{Case Study: Data Quality Issues in Rule-Based Pipelines}
\label{appendix:case_study}

Existing rule-based pipelines generate spatial QA pairs solely from scene-level metadata, without conditioning on the visual frames that a model actually observes during training. This decoupling introduces two systematic failure modes: (1)~the queried object may be \emph{absent} from the uniformly sampled frames, rendering the question unanswerable; and (2)~coarse annotation conventions may produce ground-truth labels that \emph{contradict visual common sense}. We illustrate each with a concrete example from ARKitScenes~\citep{baruch2021arkitscenes}.

\paragraph{Issue 1: Invisible objects.}
Scene \texttt{41048225} contains, according to its metadata, 1 table, 4 chairs, 2 shelves, 8 cabinets, 1 washer, 1 sink, 1 dishwasher, 1 oven, and 1 stove. The pipeline accordingly generates:

\begin{quote}
\textit{``In centimeters, what is the longest side of the dishwasher?''} \quad (Ground truth: 88\,cm)
\end{quote}

\begin{figure}[h]
  \centering
  \begin{tabular}{cccc}
    \includegraphics[width=0.22\linewidth]{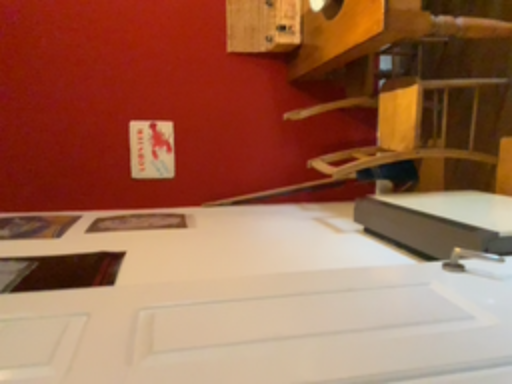} &
    \includegraphics[width=0.22\linewidth]{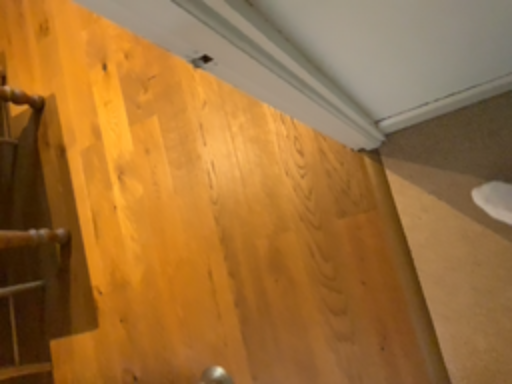} &
    \includegraphics[width=0.22\linewidth]{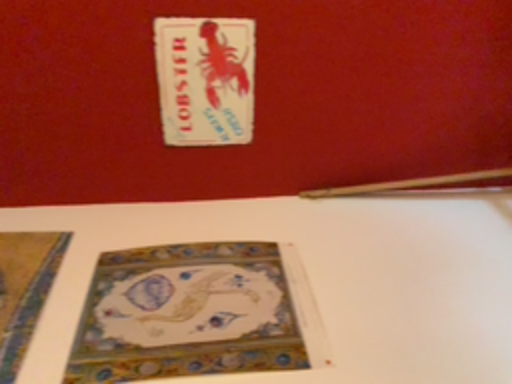} &
    \includegraphics[width=0.22\linewidth]{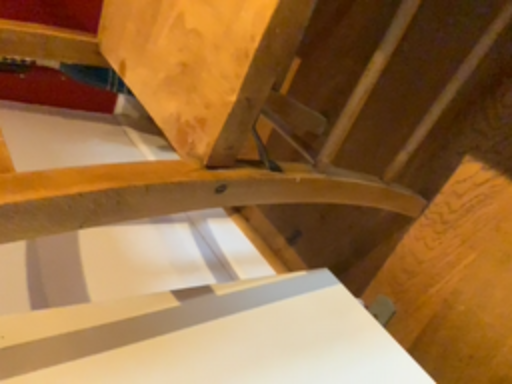} \\
    {\scriptsize 1} & {\scriptsize 3} & {\scriptsize 5} & {\scriptsize 7} \\[2pt]
    \includegraphics[width=0.22\linewidth]{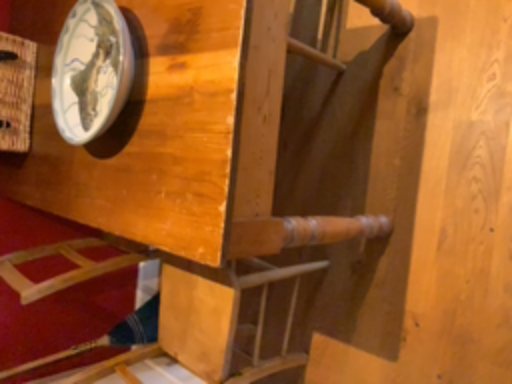} &
    \includegraphics[width=0.22\linewidth]{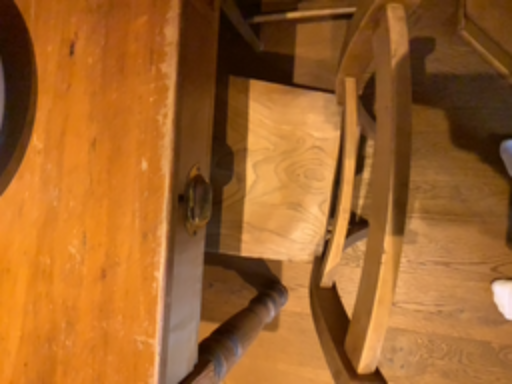} &
    \includegraphics[width=0.22\linewidth]{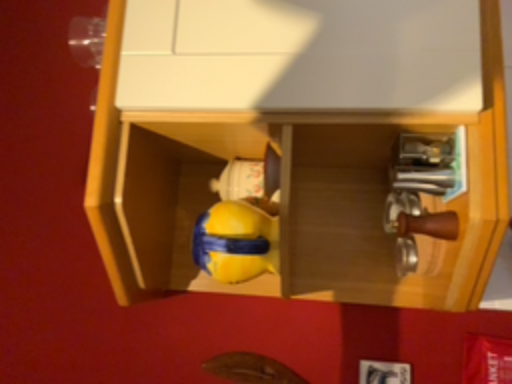} &
    \includegraphics[width=0.22\linewidth]{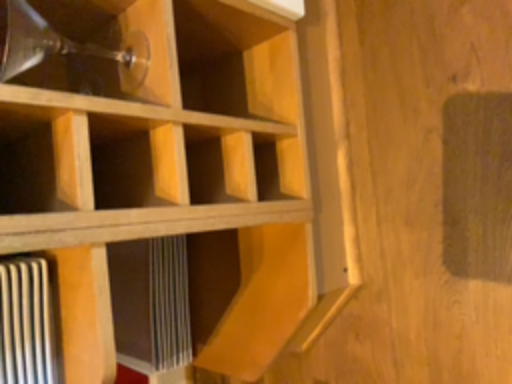} \\
    {\scriptsize 9} & {\scriptsize 11} & {\scriptsize 13} & {\scriptsize 15} \\[2pt]
    \includegraphics[width=0.22\linewidth]{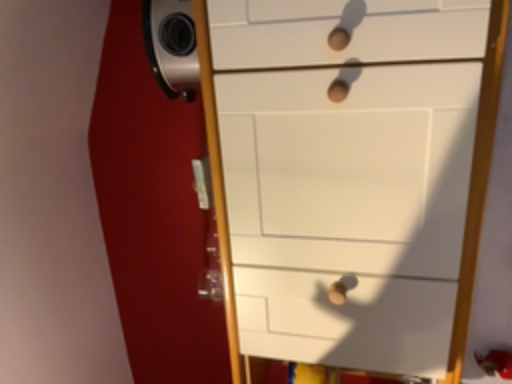} &
    \includegraphics[width=0.22\linewidth]{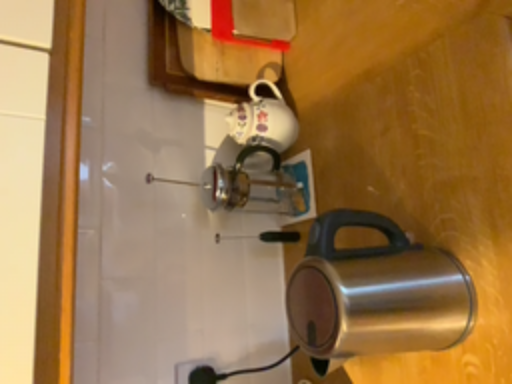} &
    \includegraphics[width=0.22\linewidth]{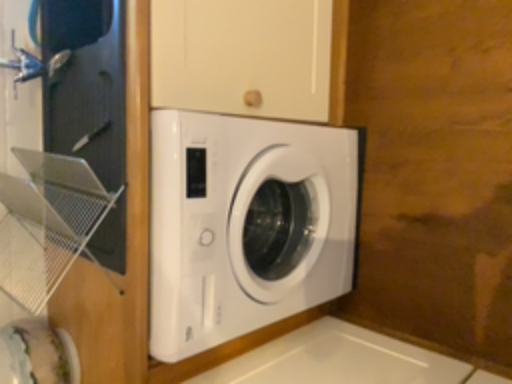} &
    \includegraphics[width=0.22\linewidth]{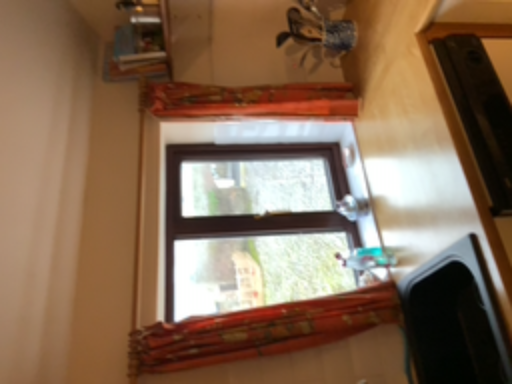} \\
    {\scriptsize 17} & {\scriptsize 19} & {\scriptsize 21} & {\scriptsize 23} \\[2pt]
    \includegraphics[width=0.22\linewidth]{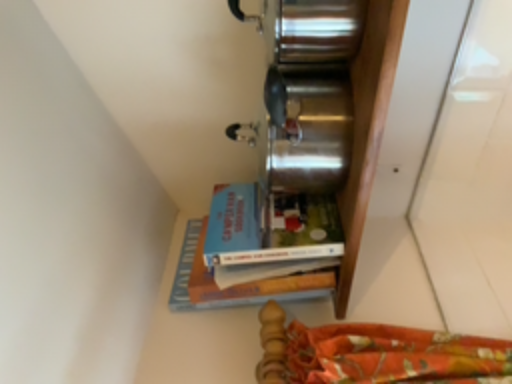} &
    \includegraphics[width=0.22\linewidth]{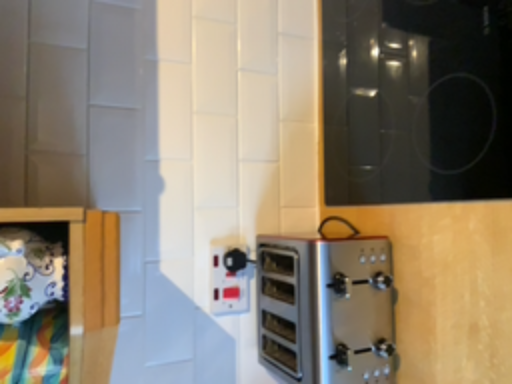} &
    \includegraphics[width=0.22\linewidth]{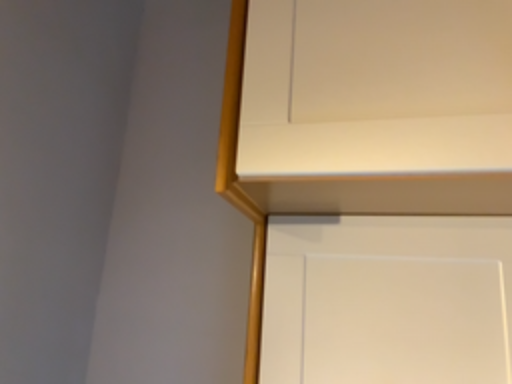} &
    \includegraphics[width=0.22\linewidth]{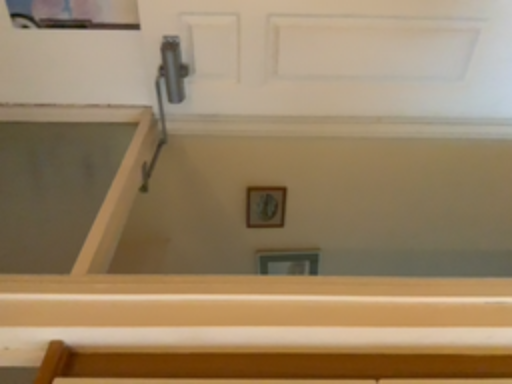} \\
    {\scriptsize 25} & {\scriptsize 27} & {\scriptsize 29} & {\scriptsize 31} \\
  \end{tabular}
  \caption{16 of the 32 uniformly sampled frames from scene \texttt{41048225} (every other frame shown). The camera covers a dining area and kitchen cabinetry. Despite the metadata listing a dishwasher, it is \textbf{absent from all 32 frames}, making the question \textit{``What is the longest side of the dishwasher?''} unanswerable from the visual input.}
  \label{fig:case_study_frames}
\end{figure}

The question is well-formed with respect to the metadata. However, the 32 uniformly sampled frames (Figure~\ref{fig:case_study_frames}) capture only a dining area and kitchen cabinetry; \textbf{the dishwasher never appears}. The model is thus supervised to answer a question for which its visual input provides no evidence, effectively being trained to hallucinate.

\paragraph{Issue 2: Annotation--visual mismatch.}
Scene \texttt{41048093} is a living room (37.1\,m\textsuperscript{2}) whose metadata records 1~fireplace, 3~sofas, 2~tables, 1~shelf, and 1~cabinet. The pipeline generates:

\begin{quote}
\textit{``How many sofas can you find in this area?''} \quad (Ground truth: 3)
\end{quote}

\begin{figure}[h]
  \centering
  \begin{tabular}{cccc}
    \includegraphics[width=0.22\linewidth]{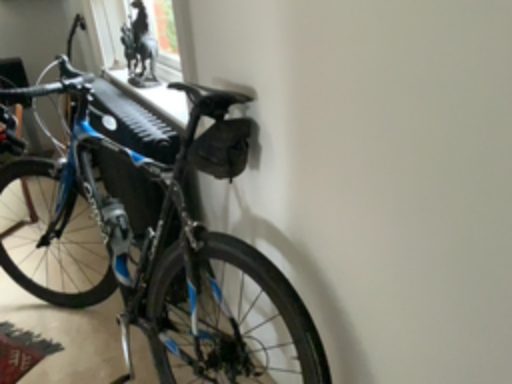} &
    \includegraphics[width=0.22\linewidth]{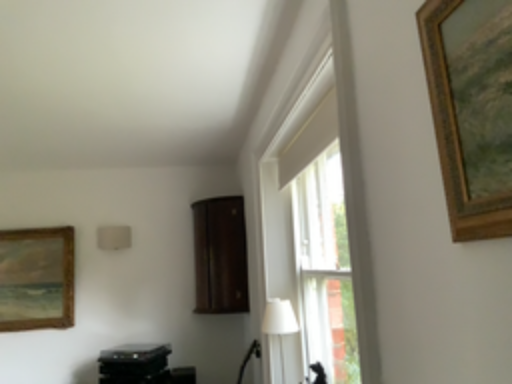} &
    \includegraphics[width=0.22\linewidth]{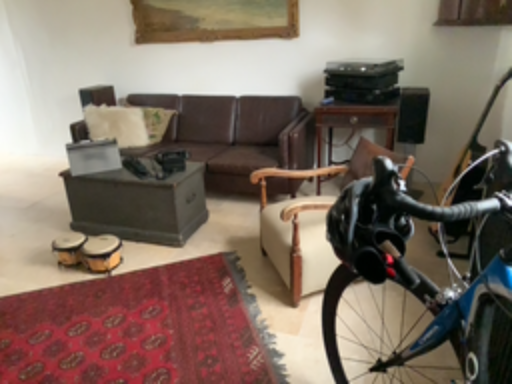} &
    \includegraphics[width=0.22\linewidth]{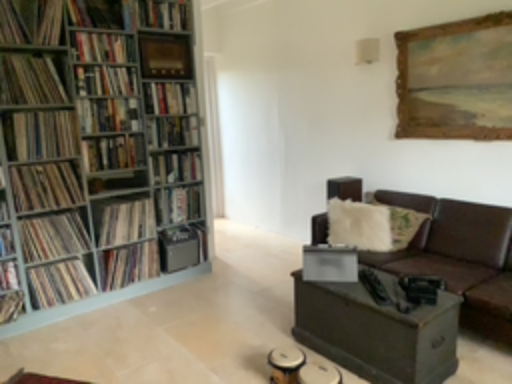} \\
    {\scriptsize 1} & {\scriptsize 3} & {\scriptsize 5} & {\scriptsize 7} \\[2pt]
    \includegraphics[width=0.22\linewidth]{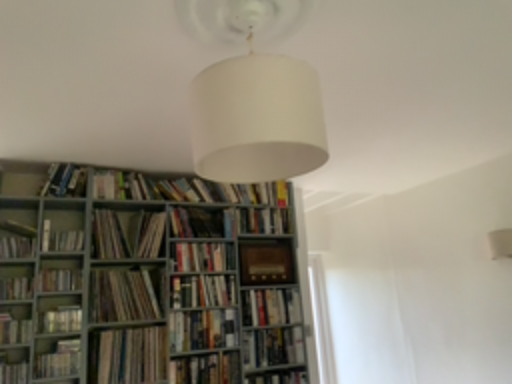} &
    \includegraphics[width=0.22\linewidth]{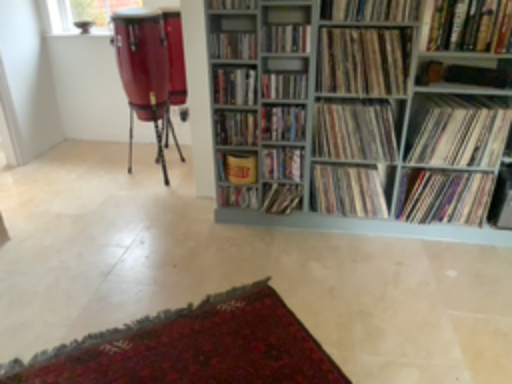} &
    \includegraphics[width=0.22\linewidth]{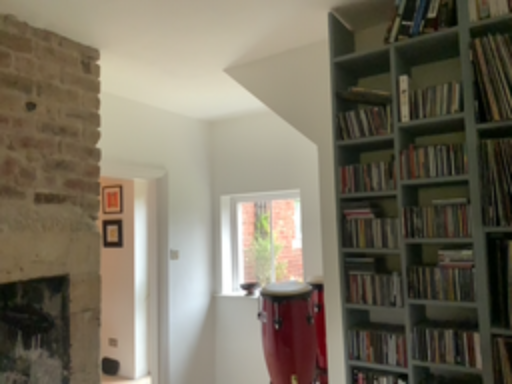} &
    \includegraphics[width=0.22\linewidth]{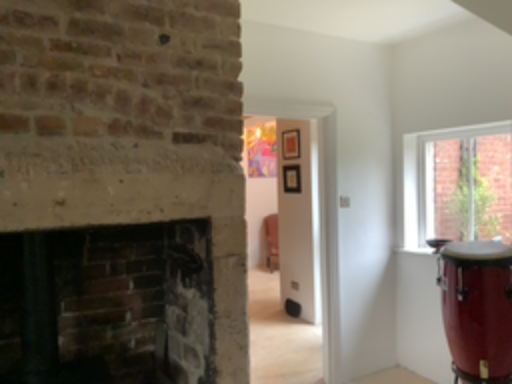} \\
    {\scriptsize 9} & {\scriptsize 11} & {\scriptsize 13} & {\scriptsize 15} \\[2pt]
    \includegraphics[width=0.22\linewidth]{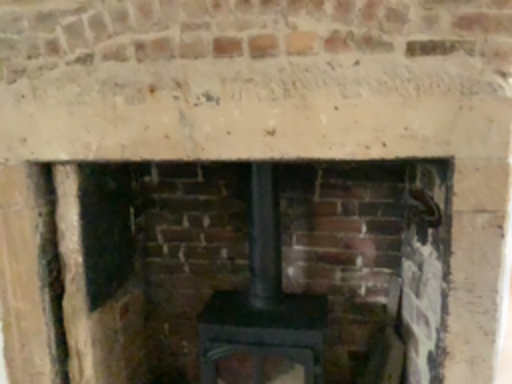} &
    \includegraphics[width=0.22\linewidth]{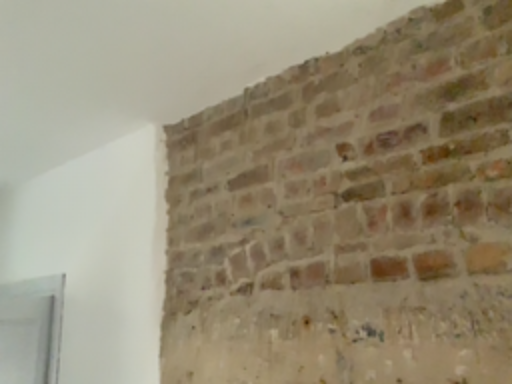} &
    \includegraphics[width=0.22\linewidth]{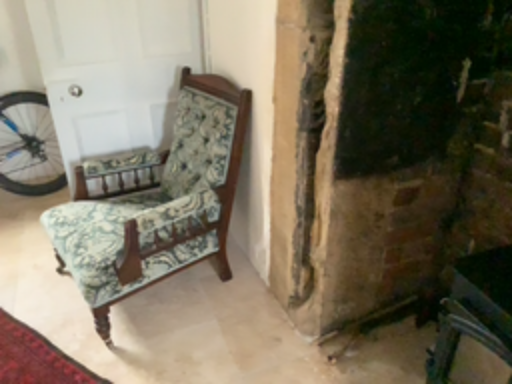} &
    \includegraphics[width=0.22\linewidth]{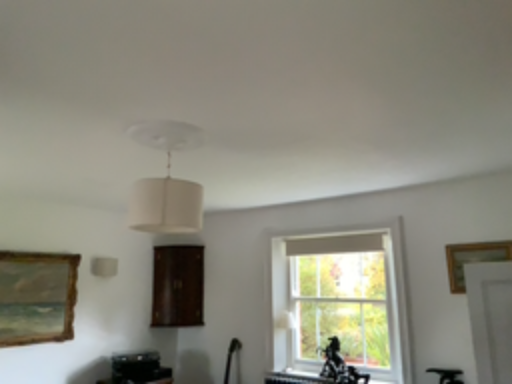} \\
    {\scriptsize 17} & {\scriptsize 19} & {\scriptsize 21} & {\scriptsize 23} \\[2pt]
    \includegraphics[width=0.22\linewidth]{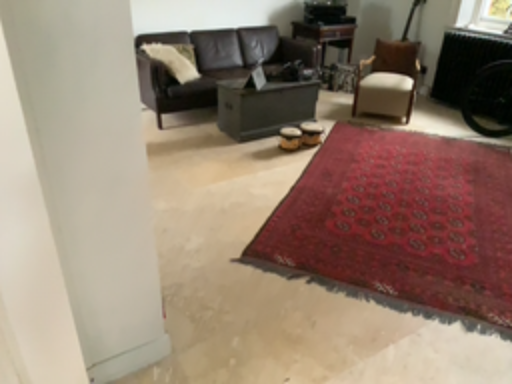} &
    \includegraphics[width=0.22\linewidth]{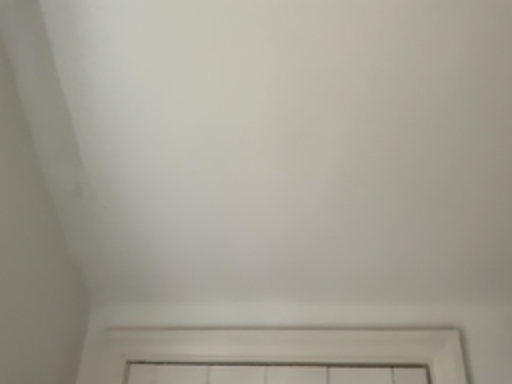} &
    \includegraphics[width=0.22\linewidth]{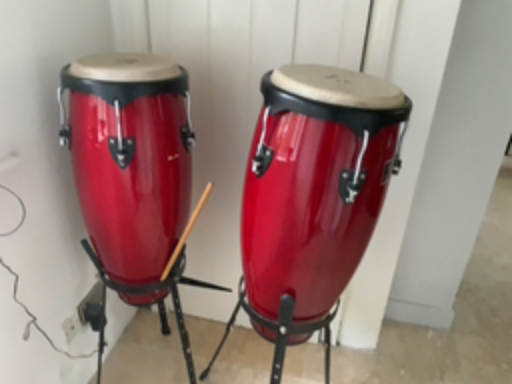} &
    \includegraphics[width=0.22\linewidth]{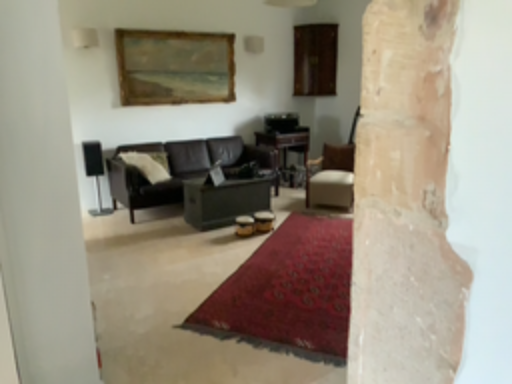} \\
    {\scriptsize 25} & {\scriptsize 27} & {\scriptsize 29} & {\scriptsize 31} \\
  \end{tabular}
  \caption{16 of the 32 uniformly sampled frames from scene \texttt{41048093} (every other frame shown). The living room contains one sofa and two armchairs, yet the 3D annotation labels all three as ``sofa,'' yielding a ground-truth count of~3. The supervised answer is \textbf{inconsistent with the visual semantics} of the scene.}
  \label{fig:case_study_sofa}
\end{figure}

Visual inspection (Figure~\ref{fig:case_study_sofa}) reveals one sofa and two armchairs around a fireplace. The 3D annotation groups all upholstered seating under the label ``sofa,'' including pieces measuring only $0.67 \times 0.57 \times 0.69$\,m---dimensions of an armchair, not a sofa. A human would not count three sofas; the ground truth reflects an annotation convention rather than visual semantics. Training on such labels teaches the model to memorize annotation artifacts instead of learning genuine visual counting.

\paragraph{Discussion.}
Both failure modes stem from the same root cause: generating questions from metadata alone, without visual grounding. Our proposer--filter pipeline (\S\ref{subsec:proposer}) addresses this by having the MLLM verify each candidate question against the sampled frames before acceptance, rejecting questions about invisible objects and labels that contradict visual evidence. Both examples above are successfully rejected by our pipeline.

\section{Broader Impact}
\label{broader impact}
Improved spatial reasoning in vision--language models can benefit applications such as
assistive navigation for visually impaired users, robotic manipulation,
and indoor scene understanding. Our self-evolving training paradigm is data-efficient and
does not require additional human annotation, reducing the cost and labor associated with
scaling spatial intelligence. However, if misused, the ability to reason about fine-grained indoor geometry could facilitate privacy violations such as unauthorized reconstruction of private spaces and advanced surveillance systems. Appropriate access controls and data governance are therefore essential when deploying such capabilities in real-world settings.

For safeguards, we have provided detailed documentation describing the model's capabilities, limitations, and intended usage. We require all users to agree to a responsible-use license before accessing model weights, which explicitly prohibits surveillance applications and unauthorized spatial reconstruction of private environments. 

\paragraph{Limitation and Future Work.}
\label{sec:limitation}

While Ouroboros-Spatial demonstrates clear advantages over existing data curation methods, several directions remain for future exploration.
(1) Learning a trainable proposer. In the current framework, the proposer is entirely driven by context engineering. Although this design is simple and stable, it limits generation diversity to what in-context learning can express. A natural extension is to train the proposer directly—e.g., via reinforcement learning—to optimize question quality, diversity, and coverage.
(2) Extending beyond annotated data. The current pipeline relies on structured scene metadata (e.g., bounding boxes, 3D positions, semantic labels) to derive verifiable ground-truth answers via executable code. This dependence restricts training to annotated 3D datasets such as ScanNet and limits applicability to in-the-wild images and videos. Developing self-supervised or reconstruction-based alternatives to metadata-dependent verification could substantially broaden the scope of the framework.
(3) Reinforcement learning for the solver. The solver is currently trained via supervised fine-tuning. While effective, incorporating reinforcement learning (e.g., GRPO) on top of the SFT checkpoint may further improve reasoning performance by optimizing sequence-level objectives aligned with downstream evaluation.